# A Convex Feature Learning Formulation
# for Latent Task Structure Discovery

**Pratik Jawanpuria**                                          PRATIK.J@CSE.IITB.AC.IN
**J. Saketha Nath**                                           SAKETH@CSE.IITB.AC.IN
Dept. of CSE, IIT Bombay, Mumbai, INDIA.

## Abstract

This paper considers the multi-task learning problem and in the setting where some relevant features could be shared across few related tasks. Most of the existing methods assume the extent to which the given tasks are related or share a common feature space to be known apriori. In real-world applications however, it is desirable to automatically discover the groups of related tasks that share a feature space. In this paper we aim at searching the exponentially large space of all possible groups of tasks that may share a feature space. The main contribution is a convex formulation that employs a graph-based regularizer and simultaneously discovers few groups of related tasks, having close-by task parameters, as well as the feature space shared within each group. The regularizer encodes an important structure among the groups of tasks leading to an efficient algorithm for solving it: if there is no feature space under which a group of tasks has close-by task parameters, then there does not exist such a feature space for any of its super-sets. An efficient active set algorithm that exploits this simplification and performs a clever search in the exponentially large space is presented. The algorithm is guaranteed to solve the proposed formulation (within some precision) in a time polynomial in the number of groups of related tasks discovered. Empirical results on benchmark datasets show that the proposed formulation achieves good generalization and outperforms state-of-the-art multi-task learning algorithms in some cases.

## 1. Introduction

The paradigm of Multi-task learning (MTL) involves learning several prediction tasks simultaneously (Caruana, 1997). In contrast to single task learning, here the idea is to synergize the related tasks by appropriate sharing of information within them. Following Evgeniou & Pontil (2004); Jacob et al. (2008); Jalali et al. (2010), tasks are said to be related if the corresponding task parameters are close to each other.

The focus of this paper is in the multi-task learning setting where some relevant features could be shared across few related tasks. Such situations arise in several real world applications (Tropp, 2006; Jalali et al., 2010). Existing works in this setting (Turlach et al., 2005; Zhang & Huang, 2008; Negahban & Wainwright, 2009; Jalali et al., 2010) employ a $\ell_1/\ell_\infty$-norm based regularizer that promotes sparsity among features and low variance among the parameters of *all* the given tasks in the shared feature space. Success of such methods depends on the extent to which the given tasks are related and the extent to which the features are shared among the tasks. In fact, Negahban & Wainwright (2009) show that $\ell_1/\ell_\infty$ regularization could actually perform worse than simple element-wise $\ell_1$ regularization when the extent to which the features are shared is less than a threshold or when the task parameters are not all close-by. Alternatively, Chen et al. (2010) assume that the relations between the tasks are known and propose employing a regularizer that penalizes deviations in weight vectors for highly correlated tasks. However, in real-world applications such task relations are not known apriori and need to be discovered.

The main contribution of this work is a convex formulation that simultaneously discovers groups of related tasks having close-by task parameters, as well as the feature space shared within each group. Here, the search space for the groups of related tasks is taken to be the power-set of the given tasks. Following the set-up of Multiple Kernel Learning (MKL) (Bach et al., 2004), the feature space in each group is taken to be that induced by a conic combination of a given set of base kernels. In the special case where the base kernels are chosen to be linear kernels formed by individual input features, this amounts to feature selection.

Note that Widmer et al. (2010) also attempt a search among all possible groups of tasks; however the runtime of their algorithm is exponential in the number of tasks. Moreover, the shared feature space in each group is assumed to be the input space. The proposed formulation employs a graph-based regularizer that encodes an important structure among the groups of tasks: if there is no feature space under which a group of tasks has close-by task parameters, then there does not exist such a feature space for any of its supersets. Note that this specialty when appropriately exploited by an algorithm may avoid search in potentially large portions of the search space which are anyway not fruitful. In Section 3, an active set algorithm is presented that exploits this specialty and optimally solves the proposed formulation in a time polynomial in the number of groups of related tasks discovered. Note that this number is typically very small compared to the size of the power-set of the given tasks. Simulations on benchmark datasets show that the proposed methodology achieves good generalization and outperforms state-of-the-art multi-task learning techniques in some cases.

The rest of the paper is organized as follows. Section 2 formalizes the notation and the problem set-up. The details of the proposed formulation and the algorithm for solving it are described in Sections 3 and 4 respectively. Experimental results are discussed in Section 5. We conclude by summarizing the work and the key contributions.

## 2. Notations and set-up

Consider a set $\mathcal{T}$ of learning tasks, $T$ in number. The training data for the $t^{th}$ task is denoted by: $\mathcal{D}_t = \{(\mathbf{x}_{ti}, y_{ti}), \ i = 1, \ldots, m\} \ \forall t = 1, \ldots, T$, where $(\mathbf{x}_{ti}, y_{ti})$ represents the $i^{th}$ input/output pair of the $t^{th}$ task. For the sake of notational simplicity, we assume that the number of training examples is the same for all the tasks. The task predictors are assumed to be affine: $F_t(\mathbf{x}) = \langle f_t, \phi(\mathbf{x}) \rangle - b_t, \ t = 1, \ldots, T$, where $f_t$ is the weight vector of the $t^{th}$ task, $\phi(\cdot)$ is the feature map and $b_t$ is the bias term. Recall that our aim is to discover groups of related tasks from the power-set of $\mathcal{T}$ (henceforth denoted by $\mathcal{V}$). To this end, we further assume that $f_t = \sum_{w \in \mathcal{G}_t} f_{tw}$ where $\mathcal{G}_t$ is the set of all subsets of $\mathcal{T}$ containing task $t$ and $f_{tw}$ is the weight vector indicating the influence of group/subset $w$ on task $t$. As we shall detail in the subsequent section, we employ a sparse regularizer that forces many $f_{tw}$ to be zero and hence enables selection of promising groups of related tasks.

In addition to discovering groups of related tasks, the proposed formulation also learns the corresponding shared feature spaces induced by conic combinations of base kernels. To this end, let $k^1, \ldots, k^n$ be the given base kernels. Let $\phi^j(\cdot)$ denote the feature map induced by the $j^{th}$ kernel $k^j, \ j = 1, \ldots, n$. Hence, $\phi(\mathbf{x}) = (\phi^1(\mathbf{x}), \ldots, \phi^n(\mathbf{x}))$. Let $f_{tw}^j$ represent the projection of $f_{tw}$ onto the $\phi^j$ space. In other words, $f_{tw} = (f_{tw}^1, \ldots, f_{tw}^n)$. With this notation, the prediction function for the task $t$ can be rewritten as $F_t(\mathbf{x}) = \langle f_t, \phi(\mathbf{x}) \rangle - b_t = \sum_{w \in \mathcal{G}_t} \langle f_{tw}, \phi(\mathbf{x}) \rangle - b_t = \sum_{w \in \mathcal{G}_t} \sum_{j=1}^n \langle f_{tw}^j, \phi^j(\mathbf{x}) \rangle - b_t$. Note that if $f_{tw}^j = 0 \ \forall t \in w$, then the feature space corresponding to the $j^{th}$ kernel is absent in the shared feature space of the group $w$. Hence learning the the task predictors or equivalently the weight vectors $f_{tw}^j$ and the bias terms $b_t$ amounts to simultaneous discovery of latent task structure as well as the corresponding shared feature spaces. In the subsequent section a novel convex formulation for learning the optimal task predictors in the current set-up is presented.

## 3. A Novel Convex Formulation

This section presents the key contribution of the paper — a convex feature learning formulation for latent task structure discovery. Following the well-establish methodology of regularized risk minimization (Vapnik, 1998), we consider the following problem:

$$\min_{f_t, b_t \ \forall t} \ \Omega(f_1, \ldots, f_T)^2 + C \sum_{t=1}^T \sum_{i=1}^m \ell(F_t(\mathbf{x}_{ti}), y_{ti}) \quad (1)$$

where $\Omega(f_1, \ldots, f_T)^2$ is the regularizer, $\ell(\cdot, \cdot)$ is a suitable convex loss function (like the hinge loss) and $C$ is the regularization parameter. In multi-task learning, it is common to choose a regularizer based on some prior knowledge about the relationship among the given tasks. For example, when all the tasks are independent, $\Omega(f_1, \ldots, f_T)^2$ can be taken as $\sum_{t=1}^T \|f_t\|_2^2$, leading to a factorization of the problem into problems involving individual tasks. In cases where it is known that all the given tasks have close-by weight vectors (i.e., all tasks are related), the following regularizer may be employed (Evgeniou & Pontil, 2004):

$$\Omega(f_1, \ldots, f_T)^2 = \mu \|h_0\|_2^2 + \sum_{t=1}^T \|h_t\|_2^2, \quad (2)$$

where $f_t = h_0 + h_t \ \forall \ t = 1, \dots, T$ and $\mu$ is the parameter that controls the trade-off between regularizing the mean weight vector $h_0$ and the variance in the weight vectors of the tasks.

In the following text, we present a novel regularizer suitable for the current problem. We begin by writing down a basic term in the proposed regularizer, $\Theta_w^j$, which induces close-by feature weights in the group $w$ wrt. the feature space $j$: $\Theta_w^j = \left( \mu \|h_{0w}^j\|_2^2 + \sum_{t \in T_w} \|h_{tw}^j\|_2^2 \right)^{\frac{1}{2}}$ where $f_{tw}^j = h_{0w}^j + h_{tw}^j$. This term is motivated from (2). Note that, $\Theta_w^j = 0 \Rightarrow f_{tw}^j = 0 \ \forall \ t \in w$ i.e., the shared feature space of the group $w$ does not involve the $j^{th}$ kernel/feature space.

Now, the terms $\Theta_w^j$, $j = 1, \dots, n$ are combined using a $p$-norm expression, leading to: $\|\Theta_w\|_p = \left( \sum_j \left( \Theta_w^j \right)^p \right)^{\frac{1}{p}}$ where $\Theta_w$ is the vector with entries as $\Theta_w^j$, $j = 1, \dots, n$, and $p \in (1, 2)$. Such a $p$-norm promotes sparsity in the selection of the kernel induced feature spaces (i.e., forces many $\Theta_w^j = 0$). With the interpretation of $\Theta_w^j$ noted above, essentially this enables feature learning within the $w^{th}$ group of tasks. Also, $\|\Theta_w\|_p = 0 \Rightarrow \Theta_w^j = 0 \ \forall \ j = 1, \dots, n \Rightarrow f_{tw}^j = 0 \ \forall \ t \in w$, $j = 1, \dots, n$ i.e., in case the node $w$ does not contain related tasks (under any feature space induced by combinations of the given base kernels), then it does not influence any of the task predictors $F_t$.

With this interpretation, one naive way of obtaining few promising groups of related tasks (that share a feature space) is by employing a $\ell_q$, $q \in (1, 2)$ norm over the terms $\|\Theta_w\|_p$, $w \in \mathcal{V}$: $\Omega(f_1, \dots, f_T) = \left( \sum_{w \in \mathcal{V}} (\|\Theta_w\|_p)^q \right)^{\frac{1}{q}}$. However the problem with this regularizer is that it renders the formulation (1) infeasible for real-world applications as the resultant optimization problem cannot be, in general, solved in a time polynomial in the number of tasks.

One key idea in the paper is to employ a graph-based regularizer, that alleviates this problem by exploiting a special structure among the groups of tasks. Note that the groups of tasks can be represented as nodes of a directed acyclic graph with the partial order $\subseteq$, representing the "subset of" relation. It can verified that $\langle \mathcal{V}, \subseteq \rangle$ is a lattice. The topmost node of the lattice represents a *dummy* node – the group with no tasks in it, the second level nodes represents groups with single task and so on. The bottommost node represents the group consisting of all the $T$ tasks. As discussed earlier, we would like to encode into our regularizer the following structure among the groups of tasks: if there is no feature space under which a group of tasks has similar task parameters, then there does not exist such a feature space for any of its supersets. In the context of the present lattice, this is same as saying: if a node $w$ is not selected, then the entire sub-lattice $D(w)$, which consists of all the descendants of $w$ (including $w$ itself), need not be selected. In the following, a regularizer that reflects this special structure is presented.

Motivated by the graph-based regularizers employed in Zhao et al. (2009); Bach (2008), we propose the following novel regularizer for the problem at hand:

$$\Omega(f_1, \dots, f_T) = \sum_{v \in \mathcal{V}} d_v \left( \sum_{w \in D(v)} \|\Theta_w\|_p^q \right)^{\frac{1}{q}} \quad (3)$$

where $q \in (1, 2)$, $p \in (1, 2)$ is a parameter that enables encoding prior knowledge regarding the task-relatedness in the group/node $v$. For e.g. one may have the prior knowledge that there is no task which is not related to the others. In this case one may choose $d_v = 0$ for all the nodes in the second level of the lattice. Note that the proposed regularizer (3) may also be viewed as a $\ell_1, \ell_q, \ell_p$ mixed-norm regularizer. The $\ell_1$-norm over the nodes ($v \in \mathcal{V}$) of the lattice promotes sparsity, and hence we have $\left( \sum_{w \in D(v)} \|\Theta_w\|_p^q \right)^{\frac{1}{q}} = 0$ for most $v \in \mathcal{V}$ i.e., few groups of related tasks are selected. Moreover, $\left( \sum_{w \in D(v)} \|\Theta_w\|_p^q \right)^{\frac{1}{q}} = 0 \Rightarrow f_{tw}^j = 0 \ \forall \ t \in w$, $\forall \ j = 1, \dots, n$, $\forall \ w \in D(v)$. In other words if a group/node is not selected (by the 1-norm), then none of its descendants are selected by the formulation — which is exactly the special structure we wanted to encode. The $q$-norm brings in additional sparsity among the descendants of the groups that are selected by the 1-norm. As we detail later, the key advantage with this regularizer is that it renders the proposed formulation (1), solvable in reasonable time.

In the following, a specialized partial dual of (1) with the proposed regularizer (3) is presented. This gives further insights into the working of the proposed formulation and motivates an efficient active set algorithm for solving it. In order to keep notations simple, the dual is presented for the case where each of the given tasks is a binary classification problem and the loss function $\ell(F_t(\mathbf{x}), y)$ is the hinge loss: $\max(0, 1 - yF_t(\mathbf{x}))$. However, it is easy to extend the derivations to other learning settings and convex loss functions as well.

**Theorem 1.** *In the case where the given tasks are all of binary classification and the hinge loss is employed as the loss function, the dual of (1) with the regularizer* defined in (3) is given by[1]

$$\min_{\gamma \in \Delta_{|\mathcal{V}|}} H(\gamma) \qquad (4)$$

where $\Delta_{|\mathcal{V}|} = \left\{ z \in \mathbb{R}^{|\mathcal{V}|} \mid z \geq 0, \ \mathbf{1}^\top z = 1 \right\}$ denotes the simplex of dimension $|\mathcal{V}|$ and $H$ is a convex function with $H(\gamma)$ equal to the optimal value of the following optimization problem:

$$\max_{\beta_t \in \mathbb{R}^{mT}} \quad \Sigma_t \mathbf{1}^\top \beta_t - \frac{1}{2} \left( \Sigma_{w \in \mathcal{V}} \lambda_w(\gamma) \left( \Sigma_{j=1}^k \left( \beta^\top \mathbf{K}_w^j \beta \right)^{\bar{p}} \right)^{\frac{\bar{q}}{\bar{p}}} \right)^{\frac{1}{\bar{q}}},$$

$$\text{s.t.} \quad \mathbf{0} \leq \beta_t \leq C\mathbf{1} \ \forall \ t, y_t^\top \beta_t = 0 \ \forall \ t, \qquad (5)$$

where $y_t$ denotes the vector with entries as $y_{ti}$, $\beta = [\beta_1 \ \dots \ \beta_T]^\top$, $\mathbf{1}$ and $\mathbf{0}$ denote vectors with all entries as $1$ and $0$ respectively, $\lambda_w(\gamma) = \left( \Sigma_{v \in A(w)} d_v^{\bar{q}} \gamma_v^{1-q} \right)^{\frac{1}{1-q}}$, $A(w)$ represents the set of ancestors for node $w$ (including $w$), $\bar{p} = \frac{p}{2(p-1)}$, $\bar{q} = \frac{q}{2(q-1)}$. The easiest way to describe the matrix $\mathbf{K}_w^j \in \mathbb{R}^{mT \times mT}$ is by writing it as a block matrix of size $T \times T$ with the $(t_1, t_2)^{th}$ block as the matrix $\mathbf{K}_w^j(t_1, t_2) \in \mathbb{R}^{m \times m}$. The $(i_1, i_2)^{th}$ entry of $\mathbf{K}_w^j(t_1, t_2)$ is =

$$\begin{cases} \frac{\mu+1}{\mu} y_{t_1 i_1} y_{t_2 i_2} k^j(\mathbf{x}_{t_1 i_1}, \mathbf{x}_{t_2 i_2}) & \text{if } t_1 = t_2 \in w, \\ \frac{1}{\mu} y_{t_1 i_1} y_{t_2 i_2} k^j(\mathbf{x}_{t_1 i_1}, \mathbf{x}_{t_2 i_2}) & \text{if } t_1, t_2 \in w, \ t_1 \neq t_2, \\ 0 & \text{otherwise.} \end{cases}$$

The dual (4) provides interesting insights into the formulation. To this end, let us begin with an interpretation for the $\mathbf{K}_w^j$ matrices. From their definition, it is easy to see that $\mathbf{K}_w^j$ can also be viewed as the gram matrix of training examples from all the tasks with an appropriately defined kernel function $k_w^j$. As $\mu \to 0$, $\frac{1}{\mu}$ dominates and the kernel function $k_w^j$ reflects great similarity between examples of tasks in $w$ (and vice-versa). Also, the examples from tasks not belonging to $w$ have low similarity with those in $w$. Hence, the kernel $k_w^j$ captures the similarity between the tasks in the group $w$ under the $j^{th}$ feature space.

Now lets focus on the problem (5). In the special case $\bar{p} = \bar{q}$, this problem is same as the $\ell_{\bar{q}}$-MKL formulation (Kloft et al., 2009) with $\hat{q} = \frac{\bar{q}}{\bar{q}-1}$ and with base kernels as $\hat{k}_w^j = (\lambda_w(\gamma))^{\frac{1}{\bar{q}}} k_w^j \ \forall w \in \mathcal{V}$ and $\forall j = 1, \dots, n$. Hence the problem (5) realizes a sparse combination of these kernels. With the interpretation provided above, this essentially amounts to a sparse selection of groups of related tasks. Hence the problem of latent task structure discovery essentially is posed as an MKL problem (with appropriately defined kernels $k_w^j$). However, unlike $\ell_{\bar{q}}$-MKL that performs a "flat" kernel selection, here the kernels are weighted by a monotonic function of $\lambda_w(\gamma)$ giving rise to a structured selection among the kernels. To see this, let us shift our focus to the dual problem (4). Because of the simplex constraints (i.e., $\ell_1$ regularization), most of the $\gamma_v$ will either be near zero at optimality. From the definition of $\lambda_w(\gamma)$, we obtain: $\gamma_v = 0 \Rightarrow \lambda_w(\gamma) = 0 \ \forall \ w \in D(v)$. Also, $\lambda_w(\gamma) = 0$ implies the entire set of kernels $k_w^1, \dots, k_w^n$ are not selected i.e., the group of tasks in $w$ are not related in any feature space under consideration. To summarize, few groups of related tasks that share a feature space are selected and if a group of tasks is unrelated ($\gamma_v = 0$), then all its supersets are unrelated (because, $\lambda_w(\gamma) = 0 \ \forall \ w \in D(v)$). Figure 1 provides an illustration of the dual problem for the case of three tasks ($T = 3$) and three base kernels ($n = 3$). The proposed formulation is equivalent to performing a structured selection among $n \times |\mathcal{V}|$ kernels arranged on a lattice. At each node, there are $n$ kernels that represent the relatedness of tasks in that group. The subsequent section, presents an efficient active-set algorithm for solving the proposed formulation that exploits the special structure in the solution described above.

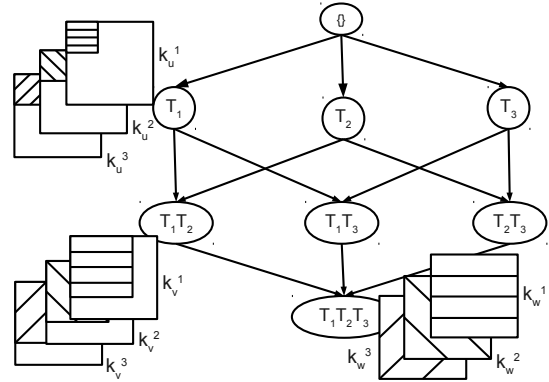

Figure 1. Figure illustrating the arrangement of kernels ($k_w^j$) selected by the dual (4). 3 tasks, $\mathcal{T} = \{T_1, T_2, T_3\}$, and 3 base kernels are considered. The kernels $k_w^j$ are shown for 3 nodes $\{T_1\}, \{T_1, T_2\}, \{T_1, T_2, T_3\}$ in the lattice. Note that these matrices have a block structure with zeros for the entries corresponding to the tasks absent in the group.

## 4. Active-set Algorithm

Following a common practice for solving large-scale convex sparsity problems (Lee et al., 2007; Bach, 2008), we propose solving the dual (4) using an active set algorithm.

The basic idea of the active set algorithm is as follows: the formulation is solved iteratively using improved guesses for the active set, which is defined as the set of $w$ for which $\gamma_w \neq 0$ at optimality. At each iteration the problem restricted to the variables in the active set is solved using an appropriate solver. In order to save computational cost, the size of the initial active set is usually taken to be minimal. After solving the problem with the variables restricted to the current active set, a sufficiency condition for optimality of the solution is verified. In case the solution is optimal, the algorithm terminates. In case it is not, the active set is updated and the restricted problem with the new active set is solved. This process is repeated until optimality is reached. Any prior knowledge related to the structure of the optimal solution may also be incorporated in building the active set at each iteration. In case of problems like (4) with sparse solutions, the hope is that one may not solve the problem with all the variables, $\gamma_w$, which are exponential in $T$ in number.

In order to formalize the active-set algorithm, we need i) an initial guess for the active set and a procedure for building/improving the active set after each iteration, ii) a sufficiency condition for verifying optimality of the current solution. Ideally, the complexity of verification of the condition should depend on the active set size rather than the problem size iii) an efficient algorithm for solving the formulation restricted to active set.

We begin with the first. Let $\mathcal{W}$ represent the active set. The initial guess as well as the methodology for the identification of the promising nodes is motivated by the special structure in the solution of (4): if a node $w$ is not selected in the optimal solution, i.e., $\gamma_w = 0$, none of the descendants of $w$ are selected (since $\lambda_v(\gamma) = 0 \ \forall v \in D(w)$). Equivalently, it can be stated that a node $w$ can be selected only if *all* its ancestors are selected i.e., only if $\gamma_v \neq 0 \ \forall v \in A(w)$. Due to this observation, $\mathcal{W}$ is always maintained to be equal to its hull, where $hull(\mathcal{W})$ is defined to be the set of all the ancestors of the nodes in $\mathcal{W}$. Accordingly, the initial guess for $\mathcal{W}$ is taken to be the second level nodes i.e., the singleton task groups. Also, in the subsequent iterations, only those nodes which have all their parents in $\mathcal{W}$ are considered as potential candidates for entry inside $\mathcal{W}$.

Towards the second requirement, we present the following theorem[2]:

**Theorem 2.** *For a given active set $\mathcal{W}$ such that $\mathcal{W} = hull(\mathcal{W})$, let the optimal solution of (4) restricted to $\mathcal{W}$ be $(\hat{\gamma}, \hat{\beta})$. Let $\hat{\Theta}$ be the value of the* term $\left( \sum_{w \in \mathcal{V}} \hat{\lambda}_w(\hat{\gamma}) \left( \sum_{j=1}^{k} \left( \hat{\beta}^{\top} \mathbf{K}_w^j \hat{\beta} \right)^{\bar{p}} \right)^{\frac{q}{\bar{p}}} \right)^{\frac{1}{q}}$. *Then,* $(\hat{\gamma}, \hat{\beta})$ *is an optimal solution of (4) with duality gap $\epsilon$ if:*

$$\max_{s \in sources(\mathcal{W}^c)} \left( \sum_{w \in D(s)} \frac{\hat{\beta}^{\top} \left( \sum_{j} \mathbf{K}_w^j \right) \hat{\beta}}{\left( \sum_{v \in A(w) \cap D(s)} d_v \right)^2} \right) \leq \hat{\Theta} + 2\epsilon \tag{6}$$

*where $sources(\mathcal{W})$ is the set of nodes in $\mathcal{W}$ with no parent in $\mathcal{W}$ and $\mathcal{W}^c$ denotes the set of all the nodes present in the lattice $\mathcal{V}$ but not in $\mathcal{W}$.*

As mentioned earlier, the above sufficiency condition is useful only if it can be verified in polynomial time in $|\mathcal{W}|$. Firstly, size of $sources(W^c)$ is upper-bounded by $T|\mathcal{W}|$. The denominator in the summation, $\sum_{v \in D(s)} d_v$ can be computed in $O(T)$ provided $d_v$ is decomposable as a product. In the simulations we use, $d_v = 1.5^{|v|}$. Because of the block structure of the matrices $\mathbf{K}_w^j$, the sum over descendants in (6) can be computed in $O(T^2 m^2)$.

In the following, we present an efficient algorithm of solving (4) restricted to $\mathcal{W}$. Note that (4) has a simple constraint set, which is a simplex and the gradient $\nabla H(\gamma)$ can be computed using the Danskin's theorem (Bertsekas, 1999): the $i^{th}$ component of this sub-gradient is given by $(\nabla H(\gamma))_i = -\frac{d_i^q \gamma_i^{-q}}{2\hat{q}} \times \left( \sum_{w \in \mathcal{V}} \lambda_w(\gamma) \left( \sum_{j=1}^{k} \left( \bar{\beta}^{\top} \mathbf{K}_w^j \bar{\beta} \right)^{\bar{p}} \right)^{\frac{q}{\bar{p}}} \right)^{\frac{1}{q}-1} \times \left( \sum_{w \in D(i)} \lambda_w(\gamma)^q \left( \sum_{j=1}^{k} \left( \bar{\beta}^{\top} \mathbf{K}_w^j \bar{\beta} \right)^{\bar{p}} \right)^{\frac{q}{\bar{p}}} \right)$, where $\bar{\beta}$ is an optimal solution of (5) with the given $\gamma$. Hence one can employ projected gradient-descent algorithm or any of its variants for solving (4. Here we employ the mirror-descent algorithm (Ben-Tal & Nemirovski, 2001) for solving (4)[3]. Note that the gradient computation $\nabla H(\gamma)$ requires solving (5) with the given $\gamma$. Also, since the constraint set in (5) is similar to that in an SVM, the $\bar{\beta}$ will be sparse at optimality. Hence we use a sequential minimal optimization (SMO) algorithm (Platt, 1999) for solving (5). Algorithm 1 summarizes the proposed active-set method. The computational complexity of the active set algorithm is as follows: let the final size of the active set be $W$. Hence, (4) is solved a maximum of $W$ times. Each run of mirror-descent algorithm takes $O(log(W))$ iterations (Ben-Tal & Nemirovski, 2001) while in each iteration the dominant computation is that of SMO for solving (5). A conservative complexity estimate for

---

---

**Algorithm 1** Active Set Algorithm

---

**Input:** Training data $\mathcal{D}_t \ \forall t$, tolerance $\epsilon$.
**Output:** $\gamma, \beta, \mathcal{W}$.
Initialize $\mathcal{W} = \{w \mid w \in \mathcal{V}, \ |T_w| = 1\}$ (i.e. first level nodes of the lattice $\mathcal{V}$)
**repeat**
    For the current $\mathcal{W}$, solve for $\gamma, \beta$ in (4) using mirror-descent & SMO.
    Calculate $V =$ nodes violating the condition (6)
    Update $\mathcal{W} = \mathcal{W} \cup V$
**until** V is empty

---

the SMO algorithm is $O((Tm)^3(Wn)^2)$. The computing cost for kernel matrices is $O(n(Tm)^2)$, while that of verifying the sufficiency condition is $O((Tm)^2 TW)$. Thus the overall complexity is: $O(n^2 W^2 m^3 T^3)$.

We end this section by presenting a variant of the proposed methodology. The motivation for this variant comes from a closer-look at the complexity of the active-set algorithm. Since the active-set always satisfies the condition $\mathcal{W} = hull(\mathcal{W})$ and since the complexity depends on the active-set size $W$; in practice one cannot realize situations where the group selected is way down the lattice. In other words, it is rare that a group with large number of tasks is selected. However, as shown in simulations, there might exist applications where weight vectors are extremely close-by for all or most of the tasks. Hence realizing a group containing most of the tasks may be beneficial. One simple modification of the proposed methodology for selecting such large groups is: invert the lattice of groups of tasks i.e., revert the parent-child relations, and employ exactly the same formulation (the descendants become the ancestors and vice-versa). It is easy to see that in this case groups involving large number of tasks may be selected; whereas selecting groups involving few tasks is now improbable. Though this modification is simple and interesting, the natural motivation for employing the graph-based regularizer is absent in this case. The graph-based regularizer needs to be motivated purely from a computational perspective in this case.

## 5. Experimental Results

In this section we present our empirical studies on the following benchmark multi-task classification and regression datasets:

**Sarcos** A multiple-output regression dataset used in Zhang & Yeung (2009). The aim is to predict inverse dynamics corresponding to the seven degrees-of-freedom of a robot arm. The number of tasks is 7 and there are 21 real valued input features. Following Zhang & Yeung (2009), we sampled 2000 random examples from each task.

**Parkinson** A multi-task regression dataset[4]. The aim is to predict Parkinson's disease symptom score for patients at different times using 19 bio-medical features. The dataset has 5,875 observations for 42 patients. The symptom score prediction problem for each patient is considered as a regression task. Thus there are 42 regression tasks with number of instances for each task ranging from 101 to 168.

**Yale** A face recognition dataset from Yale face base[5]. It contains 165 images of 15 subjects. Following the experimental setup in Zhang & Schneider (2010), each task is defined as the binary classification problem of classifying two subjects. Thus there are 28 tasks and the number of features is 30.

**Landmine** A benchmark multi-task classification dataset used in Xue et al. (2007); Zhang & Schneider (2010). It contains examples collected from various landmine fields. Each example is represented as a 9-dimensional real valued feature vector. Each task is a binary classification problem with the goal being to predict landmines (positive class) or clutter (negative class). Following Xue et al. (2007); Zhang & Schneider (2010), we jointly learn 19 tasks from the landmine fields numbered $1 - 10$ and $16 - 24$ in the data set. Number of instances in each task varies from 445 to 690. The dataset is highly biased against the positive class.

**MHC-I** A multi-task classification dataset used in Jacob et al. (2008). It contains binding affinities of various peptides with different MHC-I molecules. Each task here is a binary classification problem. We perform experiments on the same 10 tasks reported in Jacob et al. (2008). Total number of instances in the 10 tasks is 1200 and the input space consists of 180 binary features. The number of instances per task varies from 59 to 197 and the the dataset is biased against the positive class.

**Letter** A multi-task classification dataset used in Ji & Ye (2009). It consists of handwritten letters from different writers. Each task is a binary classification problem of distinguishing between pairs of letters. There are 9 such binary classification tasks and we randomly sampled 300 data points per task for our simulations. The input features used are the $8 \times 16 = 128$ binary pixels.

Each dataset was further randomly split into training and test sets. In Landmine, MHC-I and Letter datasets, random 20%-80% train-test splits were con-

---

*Table 1.* Performance of various methods on regression and classification datasets

| Dataset | STL | MTL | CMTL | DMTL | MTFL |
|---------|-----|-----|------|------|------|
| | | | Regression Datasets – Explained Variance (%) | | |
| SARCOS | $40.47 \pm 7.56 (< 1m)$ | $34.50 \pm 10.19 (< 1m)$ | $33.02 \pm 13.42 (< 1m)$ | $40.59 \pm 10.24 (< 1m)$ | $\mathbf{49.86 \pm 6.34^*} (2m)$ |
| PARKINSON | $2.84 \pm 7.51 (< 1m)$ | $4.94 \pm 19.95 (< 1m)$ | $2.74 \pm 3.62 (< 1m)$ | $-12.25 \pm 7.41 (< 1m)$ | $\mathbf{16.79 \pm 10.81^*} (23m)$ |
| | | | Classification Datasets – AUC (%) | | |
| YALE | $93.36 \pm 2.33 (< 1m)$ | $96.35 \pm 1.64 (< 1m)$ | $95.20 \pm 2.12 (< 1m)$ | $92.44 \pm 2.81 (9m)$ | $\mathbf{96.98 \pm 1.55^*} (18m)$ |
| LANDMINE | $74.60 \pm 1.55 (< 1m)$ | $76.42 \pm 0.78 (1m)$ | $75.86 \pm 0.66 (< 1m)$ | $65.86 \pm 2.63 (< 1m)$ | $\mathbf{76.44 \pm 0.92} (14m)$ |
| MHC-I | $69.25 \pm 2.07 (< 1m)$ | $72.28 \pm 1.94 (< 1m)$ | $\mathbf{72.56 \pm 1.36^*} (< 1m)$ | $58.39 \pm 5.46 (4m)$ | $71.68 \pm 2.20 (15m)$ |
| LETTER | $\mathbf{61.24 \pm 0.82} (< 1m)$ | $61.02 \pm 1.56 (< 1m)$ | $60.52 \pm 1.09 (< 1m)$ | $59.34 \pm 1.39 (< 1m)$ | $60.45 \pm 1.75 (12m)$ |

sidered. In the case of Sarcos dataset 15 random samples per task were used for training and the rest for testing. For Parkinson dataset, 5 random examples per task were used in training and the rest for testing.

We compare the following multi-task learning techniques in terms of generalization ability:

**MTL** Classical multi-task learning algorithm by Evgeniou & Pontil (2004). Assumes that all tasks are related and have close-by weight vectors. No feature learning is performed.

**CMTL** The clustered multi-task learning formulation proposed in Jacob et al. (2008). Forms clusters of tasks having similar weight vectors. No feature learning is performed.[6]

**DMTL** The multi-task feature learning formulation in Jalali et al. (2010). Performs feature selection to discover features shared across all the tasks as well as task-specific features. Also, induces close-by weight vectors for the tasks in the shared feature space.[7]

**MTFL** The proposed multi-task feature learning formulation. The base kernels were taken as linear kernels with individual input features. In addition, the linear kernel using all input features was also employed as a base kernel. Thus if the input space dimensionality is $n$, then we generate $n + 1$ linear kernels. We did not employ non-linear kernels for the sake of being fair and comparable to DMTL. The parameters $p, q$ are both fixed at 1.5, promoting sparsity in selecting groups of related tasks as well as in selecting the kernel induced feature spaces. Since the base kernels include individual input-feature based linear kernels, this amounts to feature selection.[8]

**STL** A baseline approach in which the tasks are learned independently using SVM.

Note that all of the above multi-task learning techniques rely on the same notion of task-relatedness: weight vectors of related tasks are close. Hence a comparison among them is indeed meaningful.

The free parameters in all the methods were tuned using nested 3-fold cross validation procedure. The details of the parameter ranges are as follows: in case of **MTFL**, **MTL** and **STL**, the regularization parameter $C$ was chosen from the set $\{10^{-3}, 10^{-2}, \ldots, 10^3\}$. **MTFL** and **MTL** have an additional parameter $\mu$, which was chosen from the set: $\{10^{-3}, 10^{-2}, \ldots, 10\}$. **CMTL** has 4 parameters and we considered 3 values for each leading to $3^4 = 81$ combinations (Zhang & Schneider, 2010). **DMTL** has 2 parameters and we considered 7 values of each leading to 49 combinations.

Results of the simulations are summarized in Table 1. In case of regression datasets we report the explained variance, whereas for classification datasets we report AUC (Area Under Curve). In both cases, higher the value reported, the better the algorithm. Also, we report both the mean and standard deviation in the values over 10 random train-test splits. The numbers in the brackets indicate the run-times in minutes with the tuned parameters on a Xeon machine with 16GB RAM. The best result in each dataset is highlighted. In case the best result is with the proposed method (**MTFL**) and its improvement over state-of-the-art is statistically significant, then we additionally mark it with a '*'. In case the best result is with an existing method and its improvement over the proposed method (**MTFL**) is statistically significant, then we again mark it with a '*'. Statistical significance test is performed using the paired t-test at 90% confidence.

The proposed method outperformed state-of-the-art in both the regression datasets and achieved significant improvement in case of the Yale dataset. Note that in case of Sarcos dataset, the baseline **STL** performs bet-ter than **MTL** showing that there may be some tasks that are not related to some others and the task structure is non-trivial. Hence discovering the latent task structure is indeed important in this case. The excellent performance of the proposed method indicates that the task structure is well discovered by it.

---

[6] Code available at http://cbio.ensmp.fr/~ljacob/documents/cmtl-code.tgz

[7] Code available at http://www.ali-jalali.com/index_files/L1Linf_LASSO.r

[8] Code available at www.cse.iitb.ac.in/~pratik.j/MTFL_icml12.tar.gz

According to the results, the proposed methodology does not seem to improve over state-of-the-art in case of the MHC-I and Letter datasets. A closer look at the datasets and the predictors achieved with state-of-the-art showed that the weight vectors are extremely close-by in these datasets. This motivated us to try the inverted lattice trick described towards the end of section 4. With this modified methodology we achieved an improved average AUC of 72.77% and 61.12% respectively on MHC-I and Letter datasets.

We end this section with a discussion on the run-time of the proposed method. Note that none of the existing methods attempt an optimal search over the exponentially large space of groups of tasks. Hence, as expected, the run-time of the proposed method is on the higher-side. Though this is the case, it is interesting to note that the extremely large search space ($2^{42}$) in case of the Parkinson dataset is searched in a reasonable time of 23min. Moreover, in most datasets the proposed method achieves better generalization.

## 6. Conclusions

In real-world applications it is important to discover groups of related tasks that share a feature space. The main contribution of the work is a convex formulation for solving this problem. Using a novel graph based regularizer, the search in the exponentially large space of groups of tasks is rendered feasible. Experimental results illustrate the efficacy of the proposed approach.

### Acknowledgments

We thank Ganesh Ramakrishnan for insightful discussions on this paper.

### References

Bach, F. Exploring Large Feature Spaces with Hierarchical Multiple Kernel Learning. In *NIPS*, 2008.

Bach, F., Lanckriet, G. R. G., and Jordan, M. I. Multiple Kernel Learning, Conic Duality, and the SMO Algorithm. In *ICML*, 2004.

Ben-Tal, A. and Nemirovski, A. Lectures on Modern Convex Optimization: Analysis, Algorithms and Engineering Applications. *MPS/ SIAM Series on Optimization*, 1, 2001.

Bertsekas, D. *Non-linear Programming*. Athena Scientific, 1999.

Caruana, R. Mutitask Learning. *Machine Learning*, 28: 41–75, 1997.

Chen, X., Kim, S., Lin, Q., Carbonell, J. G., and Xing, E. P. Graph-structured multi-task regression and an efficient optimization method for general fused lasso. *CoRR*, abs/1005.3579, 2010.

Evgeniou, T. and Pontil, M. Regularized multi–task learning. In *ACM SIGKDD*, 2004.

Jacob, L., Bach, F., and Vert, J. P. Clustered Multi-Task Learning: A Convex Formulation. In *NIPS*, 2008.

Jalali, A., Ravikumar, P., Sanghavi, S., and Ruan, C. A dirty model for multi-task learning. In *NIPS*, 2010.

Ji, Shuiwang and Ye, Jieping. An Accelerated Gradient Method for Trace Norm Minimization. In *ICML*, 2009.

Kloft, M., Brefeld, U., Sonnenburg, S., Laskov, P., Müller, K. R., and Zien, A. Efficient and accurate lp-norm multiple kernel learning. In *NIPS*, 2009.

Lee, H., Battle, A., Raina, R., and Ng, A. Y. Efficient sparse coding algorithms. In *NIPS*, 2007.

Negahban, S. and Wainwright, M. Phase transitions for high-dimensional joint support recovery. In *NIPS*, 2009.

Platt, J. C. *Fast training of support vector machines using sequential minimal optimization*, pp. 185–208. MIT Press, 1999.

Tropp, J. A. Algorithms for simultaneous sparse approximation: part ii: Convex relaxation. *Signal Process.*, 86: 589–602, 2006.

Turlach, B. A., Venables, W. N., and Wright, S. J. Simultaneous variable selection. *Technometrics*, 47:349–363, 2005.

Vapnik, V. *Statistical Learning Theory*. Wiley-Interscience, 1998.

Widmer, C., Toussaint, N., Altun, Y., and Ratsch, G. Inferring latent task structure for multi-task learning by multiple kernel learning. *BMC Bioinformatics*, 11:S5, 2010.

Xue, Y., Liao, X., Carin, L., and Krishnapuram, B. Multi-task learning for classification with dirichlet process priors. *Journal of Machine Learning Research*, 8:35–63, 2007.

Zhang, C. and Huang, J. The sparsity and bias of the lasso selection in high-dimensional linear regression. *Annals of Statistics*, 36:15671594, 2008.

Zhang, Y. and Yeung, D. Y. Semi-Supervised Multi-Task Regression. In *ECML/PKDD*, 2009.

Zhang, Yi and Schneider, Jeff. Learning multiple tasks with a sparse matrix-normal penalty. In *NIPS*, 2010.

Zhao, P., Rocha, G., and Yu, B. Grouped and Hierarchical Model Selection through Composite Absolute Penalties. *Annals of Statistics*, 37:3468–3497, 2009.